\documentclass{article}
\usepackage{amsthm,amsmath,amssymb}
\usepackage{mathrsfs}

\usepackage{PRIMEarxiv}
\usepackage{multirow}
\usepackage[utf8]{inputenc} 
\usepackage[T1]{fontenc}    
\usepackage{hyperref}       
\usepackage{url}            
\usepackage{booktabs}       
\usepackage{amsfonts}       
\usepackage{nicefrac}       
\usepackage{microtype}      
\usepackage{lipsum}
\usepackage{fancyhdr}       
\usepackage{graphicx}       
\graphicspath{{media/}}     
\usepackage[table]{xcolor}
\title{GlocalFuse-Depth: Fusing Transformers and
CNNs for All-day Self-supervised Monocular Depth Estimation
 }

\author{
  Zezheng Zhang, Ryan K. Y. Chan \\
  The University of Hong Kong \\
  \texttt{\{zezhengz, cky166\}@connect.hku.hk} \\
   \And
  Kenneth K. Y. Wong \\
  The University of Hong Kong, Advanced Biomedical Instrumentation Centre \\
  \texttt{kywong@eee.hku.hk} \\
}

\begin{document}
\maketitle

\begin{abstract}
In recent years, self-supervised monocular depth estimation has drawn much attention since it frees of depth annotations and achieved remarkable results on standard benchmarks. However, most of existing methods only focus on either daytime or nighttime images, thus their performance degrades on the other domain because of the large domain shift between daytime and nighttime images. To address this problem, in this paper we propose a two-branch network named GlocalFuse-Depth for self-supervised depth estimation of all-day images. The daytime and nighttime image in input image pair are fed into the two branches: CNN branch and Transformer branch, respectively, where both fine-grained details and global dependency can be efficiently captured. Besides, a novel fusion module is proposed to fuse multi-dimensional features from the two branches. Extensive experiments demonstrate that GlocalFuse-Depth achieves state-of-the-art results for all-day images on the Oxford RobotCar dataset, which proves the superiority of our method.

\end{abstract}


\section{Introduction}
Monocular depth estimation has been one of the most basic tasks in computer vision field, which is widely applied in various tasks, such as simultaneous localization and mapping (SLAM) \cite{6386103,7946260,Tateno_2017_CVPR,Liu_Song_Lyu_Diao_Wang_Liu_Zhang_2021}, augmented reality (AR) \cite{6126513} and autonomous driving \cite{Menze_2015_CVPR}. However, it is inherently an ill-posed problem: a single image could be produced from an infinite number of distinct 3D scenes, thus the depth map can be completely different. Benefiting from the advance of deep learning method and availability of large-scale training data, significant improvement has been achieved in monocular depth estimation in recent years. A variety of deep learning methods have manifested their effectiveness to recover the pixel-level dense depth map from an single image in an end-to-end manner \cite{Fu_2018_CVPR,Yin_2019_ICCV,Zhang_2019_CVPR}.

However, collecting large-scale training datasets with accurate ground-truth depth maps for supervised learning is costly, due to the large size and high prize of depth sensors (like RGB-D cameras and LIDAR). In addition, the internal error and noise  characteristics of depth sensors will also affect the learning of mapping between object appearance and their depth. Therefore, self-supervised depth estimation methods have been introduced. These methods use spatial consistency (stereo depth reconstruction) or temporal consistency (monocular video depth reconstruction)  as supervisory signal during the training process. Some of them \cite{10.1007/978-3-319-46484-8_45,Godard_2017_CVPR,Zhou_2017_CVPR,Johnston_2020_CVPR,Zhao_2020_CVPR} have achieved performance comparable to the supervised methods on widely used benchmarks, such as KITTI \cite{doi:10.1177/0278364913491297} and Cityscapes \cite{Cordts_2016_CVPR}.



Even so, most of current self-supervised depth estimation methods only handle daytime image depth estimation problem. They fail to generalize well on nighttime images owing to the large domain shift between daytime and nighttime images and not suitable for all-day tasks such as automatic driving. Depth estimation of nighttime images are challenging owing to varying illuminations and low visibility at nighttime. One solution is to apply image-to-image translation methods, such as CycleGAN \cite{Zhu_2017_ICCV}, to translate nighttime images into daytime images,  then use a pretrained daytime model to estimate depth from these translated images. Unfortunately, due to the large domain shift between the two domain, it is difficult to obtain natural daytime images, then the performance is also limited. Fig. 1 (b) demonstrate the depth estimation results of Monodepth2 \cite{Godard_2019_ICCV}, a effective self-supervised daytime depth estimation approach, on nighttime images. Fig. 1 (c) demonstrate the depth estimation results of Monodepth2 \cite{Godard_2019_ICCV} with CycleGAN translated images as input. We can see that the depth details, especially edges, are failed to be estimated due to the varying illuminations and low visibility of nighttime images.

\begin{figure}[htb]
\centering\includegraphics[width=10cm]{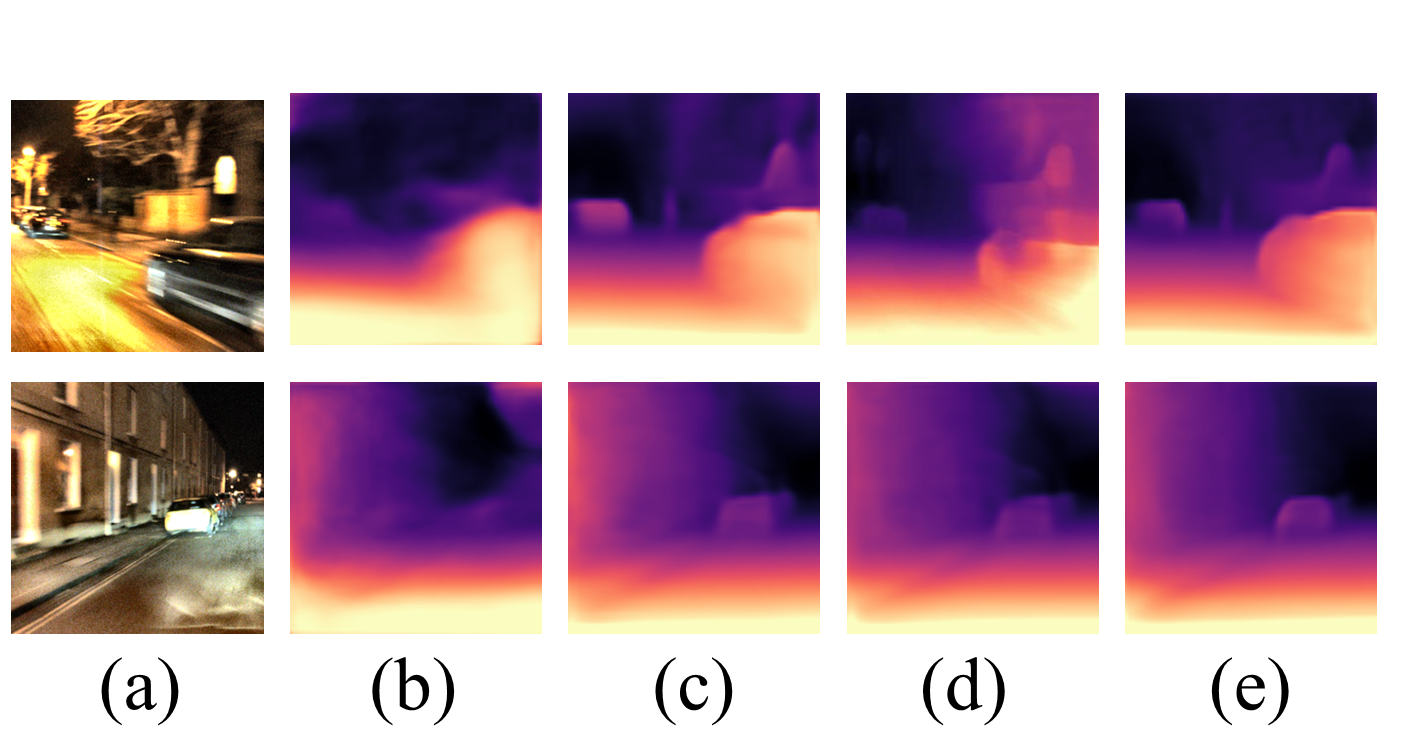}
\caption{Comparison with other methods on Oxford RobotCar dataset \cite{doi:10.1177/0278364916679498}. From left to right: (a) Nighttime Images,(b) Monodepth2 \cite{Godard_2019_ICCV}, (c) Monodepth2+CycleGAN \cite{Zhu_2017_ICCV}, (d) ADDS-DepthNet \cite{Liu_2021_ICCV}, (e) GlocalFuse-Depth.}
\end{figure}

For images captured with a fixed viewpoint, their depth map remains the same, regardless of other terms changing with time such as illumination. Also, \cite{Dijk_2019_ICCV} proves that texture information plays more important roles in depth estimation than exact color information. To cater to above issues, inspired by \cite{Liu_2021_ICCV} (ADDS-DepthNet, depth estimation results shown in Fig. 1 (d)), here we propose GlocalFuse-Depth, which adopts the day-night image pair (with same depth map, generated by CycleGAN \cite{Zhu_2017_ICCV}) as input to feed into two branches: CNN branch and Transformer branch. We choose this two-branch network architecture for the following reasons: although CNN is proved powerful at building hierarchical feature representation, its lack of efficiency in capturing global context information remains a challenge. CNN captures global information usually at the expense of efficiency, which requires stacked convolutional layers until the receptive field becomes large enough. While Transformer is good at modeling global context, it shows limitations in capturing fine-grained details due to the lack of spatial inductive-bias. In order to take full advantage of both structures, they are both adopted as two branches in GlocalFuse-Depth to capture both local and global feature (where "Glocal" comes from). Output features from the two branches with same dimension are fused to jointly make predictions. As shown in Fig. 1 (e), our method effectively relieves the problems of varying illuminations and low visibility, and achieves more attractive results for nighttime images.



The main contributions of this paper can be summarized as:

\begin{itemize}

\item We propose a two-branch network GlocalFuse-Depth for self-supervised all-day image depth estimation. It adopts complementary encoders (CNNs and Transformers) to capture local and global information in a day-night image pair to better grasp the texture correspondence between RGB image and depth map. To the best of our knowledge, GlocalFuse-Depth is the first model synthesizing CNN and Transformer for all-day image depth estimation. 

\item A new fusion module is proposed. It combines and aggregates the information from the two branches to obtain a comprehensive representation for the final depth estimation. Both channel-wise and spatial-wise attention are applied to boost representation power of our two-branch network.

\item Experimental results of our method outperform other state-of-the-art methods on the Oxford RobotCar dataset for all-day image depth estimation, which confirms the superiority of our method.

\end{itemize}


\section{Related Work}

\subsection{Daytime Depth Estimation}

In view of the shortcomings of supervised depth estimation, self-supervised depth estimation has been extensively studied in recent years. There are two main types of self-supervised methods: stereo depth reconstruction and monocular video depth reconstruction. \cite{Godard_2017_CVPR} is the pioneering work of stereo depth reconstruction. They use the left–right consistency constraints as the supervisory signal to train the depth estimation network. On the basis, other researchers make a series of improvements on this work. For example, \cite{Wong_2019_CVPR} introduces an adaptive regularization scheme for the network to better handle the co-visible and occluded regions in a stereo pair. \cite{8885578} proposes a sparsity-invariant autoencoder to improve the traditional visual odometry. On the other hand, \cite{Zhou_2017_CVPR} is the first self-supervised depth estimation method using geometry cues between monocular video frames as supervisory signal, which trains the depth network along with a separate pose network. This work provided many useful references for subsequent works. \cite{Mahjourian_2018_CVPR} trains their network with a 3D-based loss under the supervision of geometry cues from adjacent frames of monocular video. \cite{Johnston_2020_CVPR} incorporates self-attention and discrete disparity prediction into their network. These methods have achieved promising results on daytime benchmarks, such as KITTI \cite{doi:10.1177/0278364913491297} and Cityscapes \cite{Cordts_2016_CVPR}. However, self-supervised depth estimation approaches for all-day images have not been well addressed due to the challenge of varying illuminations and low visibility of nighttime images, and the performance of most current methods degrades a lot on nighttime images due to large domain shift between daytime and nighttime images.

\subsection{Nighttime Depth Estimation}


Self-supervised depth estimation for nighttime images is still a relatively underexplored topic even with today's complete deep learning method. Different approaches have also been proposed to address this challenging task from both hardware and software side. \cite{Kim_Choi_Hwang_Kweon_2018} and \cite{Lu_2021_WACV} benefit from thermal images acquired by additional sensors. \cite{Spencer_2020_CVPR} is proposed to simultaneously learn cross-domain feature representation and depth estimation to acquire more robust supervision for nighttime images. \cite{9320347} utilizes a translation network which renders nighttime stereo images from daytime stereo images. \cite{10.1007/978-3-030-58604-1_27} adapts a network trained on daytime images to work for nighttime images, aiming to transfer knowledge from daytime images to nighttime images. \cite{Wang_2021_ICCV} propose Priors-Based Regularization to learn distribution knowledge from unpaired depth maps and Mapping-Consistent Image Enhancement module to enhance image visibility and contrast. \cite{Liu_2021_ICCV} propose a domain-separated framework which partition the information of day-night image pairs into two complementary sub-spaces and relieve the influence of disturbing terms for all-day depth estimation.

Though remarkable progress has been achieved, the large domain shift between daytime and nighttime images is always hard to be fixed. It is difficult to obtain satisfactory depth map of daytime and nighttime images in one single network. To solve this problem, we propose the two branch network GlocalFuse-Depth for for all-day self-supervised depth estimation, which can effectively issue the problem of inherent domain gap between daytime and nighttime images.




\section{Approach}

\subsection{Input Image Pair and Self-supervised Depth Estimation Objective}

For the daytime and nighttime images of the same scene, the depth information should be consistent, although their light conditions are quite different. However, it is almost impossible to guarantee a real-world daytime and nighttime image pair contain identical objects even they were captured at the same position with a fixed viewpoint, since there are always moving objects in outdoor scenes. This will mislead the network when mapping the RGB information with the depth information. Therefore, a pretrained CycleGAN \cite{Zhou_2017_CVPR} is used to translate daytime images to nighttime images before training, then a daytime and the corresponding nighttime image compose a day-night image pair. This pre-processing method ensures that the two images in an image pair contain identical depth information, and they can be fed into the two branches of our network respectively with the same supervisory signal when training. Note that other image translation methods can also be used here. During the inference process, CycleGAN \cite{Zhou_2017_CVPR} is also used to translate the input daytime or nighttime image to the other domain and form the day-night image pair for the network to estimate depth.

As for the optimization objective, similar to prior work, we formulate the self-supervised depth estimation task as a monocular video depth reconstruction problem. The goal is to minimize the photometric reprojection error \cite{Godard_2019_ICCV} at training stage. For two frames $I_{t}$ and $I_{t^{'}}$ in an image sequence, the reprojection process is:

\begin{equation}
I_{t}=KT_{t^{'}\rightarrow t}D(t^{'})K^{-1}I_{t^{'}},
\end{equation}

where $D(t^{'})$ is the depth map of the source frame $I_{t^{'}}$, $T_{t^{'}\rightarrow t}$ represents the spatial transformation from $I_{t^{'}}$ to the target frame $I_{t}$ (estimated by a pose network) and $K$ is the camera intrinsic matrix. The photometric loss can be formulated as:


\begin{equation}
L_p=L_p^d+L_p^n,
\end{equation}

\begin{equation}
L^{d/n}_p=\frac{\alpha}{2}(1-SSIM(I_{t}^{d/n},\hat{I_t}^{d/n}))+(1-\alpha)||(I_{t}^{d/n}-\hat{I_t}^{d/n}||_1,
\end{equation}

where $\hat{I}$ represents the reconstructed target image obtained by Eq. (1) and $I$ represents the original image. The superscript $d$ and $n$ represent daytime and nighttime images, respectively and $\alpha$ is empirically set as 0.85. Here we use the two frames temporally adjacent to $I_{t}$ as our source frames, \emph{i.e.} $I_{t-1}$ and $I_{t+1}$.  



\subsection{Network Architecture}

\begin{figure}[htb]
\centering\includegraphics[width=16cm]{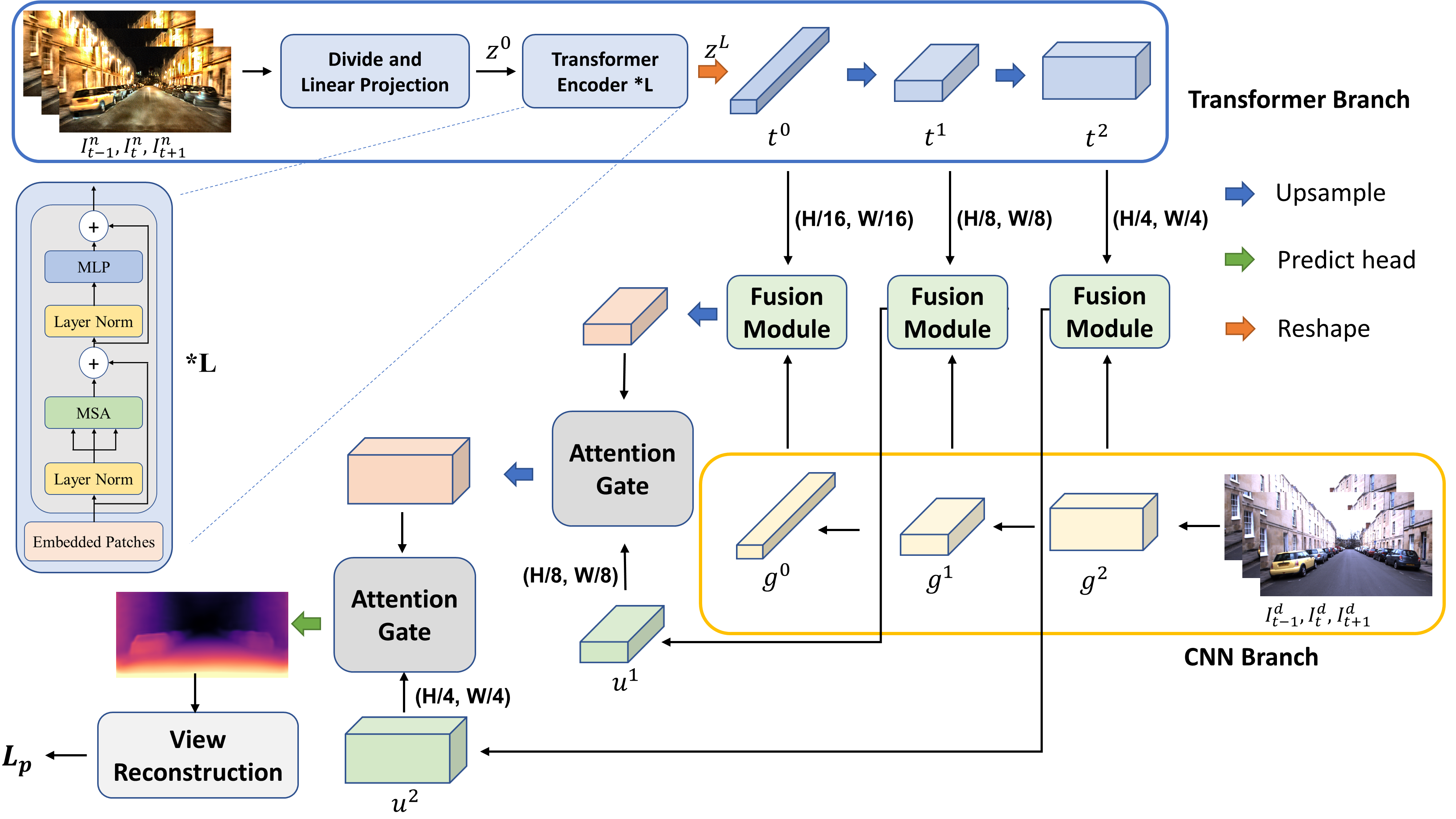}
\caption{Overview of GlocalFuse-Depth architecture: two parallel branches - CNN branch (bottom right) and Transformer branch (top) fused by our proposed fusion module.}
\end{figure}

As shown in Fig. 2, GlocalFuse-Depth consists of two parallel branches: CNN branch and Transformer branch, which takes day-night image pairs as input (daytime images fed into the CNN branch and nighttime images fed into the Transformer branch). The CNN branch gradually increases the receptive field and encodes features from local to global, and the Transformer branch starts with global self-attention and recovers the local details by upsampling. Extracted features with same spatial resolutions from two branches are fed into our proposed fusion module, where branch channel selection and spatial attention are applied to selectively fuse the information from both branches. Then the multi-level fused feature maps are combined to generate the final prediction using attention-gated skip-connection \cite{SCHLEMPER2019197}. There are two main benefits of our proposed architecture: firstly, with the characteristic of grasping long-range information of transformer, GlocalFuse-Depth can capture global depth information without stacking a large number of convolutional layers. At the same time, the use of CNN helps to preserve sensitivity of the network on local context. Secondly, our proposed fusion module comprises channel-selection and spatial attention for features from the two branches, which exploits different characteristics of the output feature from two branches, thus making the fused representation comprehensive and compact.

The forward propagation process of the Transformer branch is as follows. Firstly, the input image $x\in\mathbb{R}^{H*W*3}$ is divided into $N=\frac{H}{P}*\frac{W}{P}$ patches, where the patch size $P$ is typically set as 16. These patches are then flattened and fed into a linear embedding layer with latent size $D_0$ and obtain the embedding sequence $E\in\mathbb{R} ^{N*D_0}$. In order to utilize the spatial prior, a learnable positional embedding with the same dimension as $E$ is added. The result vector $z^0\in \mathbb{R}^{N*D_0}$ is fed into $L$ consecutive transformer encoder block. In the transformer encoder, self-attention (SA) mechanism is applied as

\begin{equation}
SA(z^i)=softmax(\frac{qk^T}{\sqrt{D_h}})v,
\end{equation}

where $[q,k,v]=z^iW_{qkv}$, $W_{qkv}\in\mathbb{R}^{D_0*3D_h} $ is the linear projection matrix and $D_h$ is the dimension of $q$, $k$ and $v$. Multi-head self-attention (MSA) is an extension of SA in which runs $k$ self-attention operations in parallel and project the concatenated outputs back to $\mathbb{R}^{D_0}$. MSA is the basis of transformer for learning long-range dependencies. Layer normalization and MLP is followed to obatin the encoded sequence $z^L\in \mathbb{R}^{N*D_0}$ (refer to \cite{https://doi.org/10.48550/arxiv.2010.11929} for other details). In the decoder part, we first reshape $z^L$ to $t^0\in\mathbb{R}^{\frac{H}{16}*\frac{W}{16}*C_0}$, then use two consecutive upsampling layer to recover its spatial resolution, obtaining $t^1\in\mathbb{R}^{\frac{H}{8}*\frac{W}{8}*C_1}$ and $t^2\in\mathbb{R}^{\frac{H}{4}*\frac{W}{4}*C_2}$. These three feature maps with different dimension are saved for late fusion with the outputs of CNN branch.

In former works, to obtain global information, CNN encoders are always designed very deep to decrease the dimension of an image, which takes up massive computational resources and increases training difficulty. Considering that global information can be grasped by the Transformer branch, here we only adopt the first 4 blocks of ResNet34 as our CNN branch. This also benefits the CNN branch from retaining richer local information. We take the outputs from the 4th ($g^0\in\mathbb{R}^{\frac{H}{16}*\frac{W}{16}*C_0}$), 3rd ($g^1\in\mathbb{R}^{\frac{H}{8}*\frac{W}{8}*C_1}$)and 2nd ($g^2\in\mathbb{R}^{\frac{H}{4}*\frac{W}{4}*C_2}$) blocks to fuse with the outputs from the Transformer branch. 
 
To generate final segmentation, fused feature with different scales are combined using the attention-gated (AG) skip-connection \cite{SCHLEMPER2019197}. Features with smaller spatial dimension are upsampled and used as the gating signal to allow the model to focus more on prominent regions while suppressing feature activations in irrelevant regions at late stage. 

\subsection{Fusion Block}
\begin{figure}[htb]
\centering\includegraphics[width=16cm]{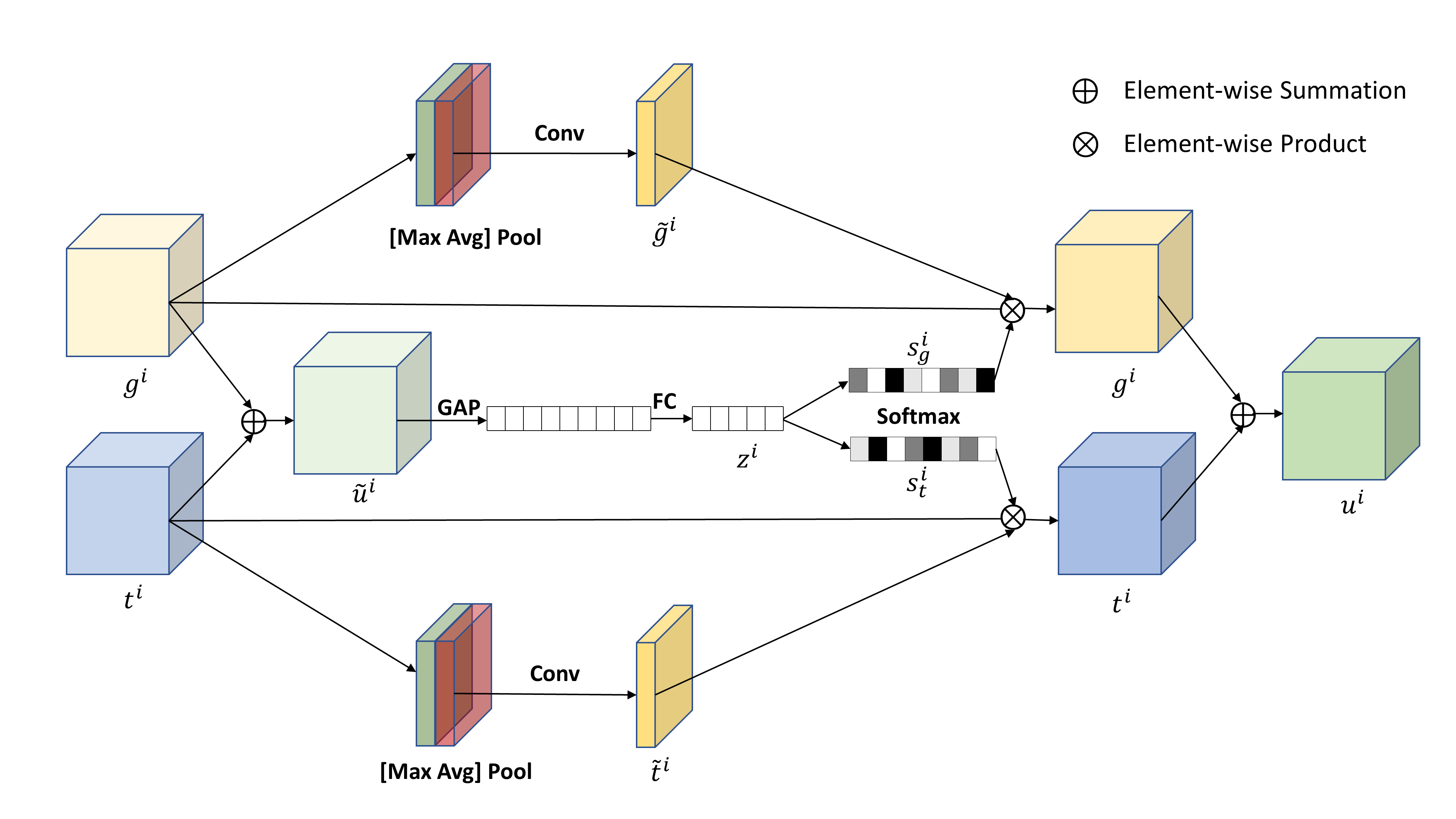}
\caption{The proposed fusion module.}
\end{figure}







To effectively combine the encoded features from the CNN branch and the Transformer branch, we propose a new fusion module (as in Fig. 3). For each scale of output of the CNN and the Transformer branch, inspired by \cite{Li_2019_CVPR}, we first select representative channel of the two outputs for the final estimation. Outputs from the two branches is aggregated via an element-wise summation: 
\begin{equation}
\hat{u}^i=t^i+g^i
\end{equation}

A gate vector $z^i$ used to control information flows from the two branches is computed by consecutive global average pooling (GAP) layer and fully connected (FC) layer.

\begin{equation}
z^i=FC(GAP(\hat{u}^i))
\end{equation}

GAP layer is applied to embed the global information of the summation feature $\hat{u_i}$ and FC layer is used for better efficiency by reducing the dimension. After that, a soft attention vector, guided by the gate vector $z^i$, is computed to select channels of the features from the two branches: 

\begin{equation}
s_t^i=\frac{e^{W_t^i*z^i}}{e^{W_t^i*z^i}+e^{W_g^i*z^i}}
\quad\mathrm \quad
s_c^i=\frac{e^{W_g^i*z^i}}{e^{W_t^i*z^i}+e^{W_g^i*z^i}}
\end{equation}

Besides, the spatial attention block adopted from CBAM \cite{Woo_2018_ECCV}, which is complementary to the soft channel attention, is used as spatial filters to highlight informative regions and suppress irrelevant regions for both branches: outputs of max pooling and average pooling along the channel axis are concatenated and fed into a convolutional layer to produce the spatial attention map:

\begin{equation}
\tilde{t}^i=Conv([Max\ Avg]Pool(t^i))
\quad\mathrm \quad
\tilde{g}^i=Conv([Max\ Avg]Pool(g^i))
\end{equation}

Then we multiply soft channel-selection vector $\tilde{t_i}$ ($\tilde{g_i}$), spatial attention vector $s_t$($s_g$) and original outputs $t_i$($g_i$) from each branch. Finally, we perform element-wise summation to get the final fuse result for each scale:

\begin{equation}
t^i=\tilde{t}^i*s_t^i*t^i
\quad\mathrm \quad
g^i=\tilde{g}^i*s_g^i*g^i
\end{equation}
\begin{equation}
u^i=t^i+g^i
\end{equation}

\section{Experiments}
In this section, we compare the performance of our method with  state-of-the-art methods on Oxford RobotCar dataset \cite{doi:10.1177/0278364916679498}, for both daytime and nighttime images.



\subsection{Dataset}

The KITTI \cite{doi:10.1177/0278364913491297} and Cityscapes \cite{Cordts_2016_CVPR} datasets are widely used in depth estimation task. However, these two datasets only consist of daytime images, which can not meet the requirements for all-day depth estimation. Therefore, we choose the Oxford RobotCar dataset \cite{doi:10.1177/0278364916679498} for training and testing in this work. Oxford RobotCar dataset is a large-scale dataset captured by cameras and sensors with fixed position and viewpoint during a long-term autonomous driving, which includes both daytime and nighttime images. Following \cite{Liu_2021_ICCV}, we use the left images collected by the front stereo-camera (Bumblebee XB3) with the resolution of 960 * 1280 for self-supervised depth estimation. Sequence ”2014-12-09-13-21-02” and ”2014-12-16-18-44-24” are used for training of daytime and nighttime, respectively. Both training data are selected from the first 5 splits. The testing images are collected from the other splits of the Oxford RobotCar dataset, which contains 451 daytime images and 411 nighttime images. We use the depth data captured by the front LMS-151 depth sensors as the ground truth in the testing phase. The images are first center-cropped to 640 * 1280, then resized to 256 * 512 as the inputs of the network.

\subsection{Implementation Details}

At the training phase, the pretrained CycleGAN\cite{Zhu_2017_ICCV} is used to translate daytime images to nighttime images. The generated 'fake' nighttime images and the corresponding daytime images compose the image pairs and fed into GlocalFuse-Depth.

The network is trained with an NVIDIA GeForce RTX3090 GPU for 30 epochs. The parameters are optimized with Adam optimizer ($\beta_1=0.9$, $\beta_2=0.99$) and the batch size is set as 16. We also used learning rate warmup scheme to increase the training stability: the learning rate increases to 1e-5 linearly in first 5 epoches, then decreases as the cosine function in remaining epoches.





\subsection{Quantitative Results}


Table. 1 demonstrates the quantitative comparison results between our method and some state-of-the-art methods. The maximum depth is set to be 60m. In Table. 1, Monodepth2 \cite{Godard_2019_ICCV} (day) and Monodepth2 \cite{Godard_2019_ICCV} (night) mean training with daytime and nighttime images of the Oxford dataset, respectively. Monodepth2+CycleGAN \cite{Zhu_2017_ICCV} means training the Monodepth2 with daytime images, then using 'fake' daytime images, which are translated from the nighttime images, to test its performance. Monodepth2 \cite{Godard_2019_ICCV} is proved to be an effective self-supervised depth estimation method for daytime images as shown in Table. 1. However, its performances are limited for nighttime images because of the large domain shift between daytime and nighttime images. Besides, because of varying illumination and the low visibility, different degrees of information is lost due to too bright or too dark regions on nighttime images, which causes that training directly on nighttime images also cannot get good enough results.

Meanwhile, although combining Monodepth2 \cite{Godard_2019_ICCV} and CycleGAN \cite{Zhu_2017_ICCV} could transform the nighttime images to 'fake' daytime images and reduce the domain shift between train and test images, the performances are also limited due to the loss of CycleGAN itself. ADFA \cite{10.1007/978-3-030-58604-1_27} and RNW-Net \cite{Wang_2021_ICCV}, which is specialized to estimate depth of nighttime images, could reduce the domain shift between day and night images at the feature level, but the performance is limited by daytime results. ADDS-DepthNet \cite{Liu_2021_ICCV} also use day-night image pair as input and could improve the depth estimation results of both the daytime and nighttime images to a certain extent, its performance can still be improved. As shown in Table. 1, the use of complementary encoders and effective fuse module of our GlocalFuse-Depth could grasp inherent texture features from both daytime and nighttime images, then improve the depth estimation performance for all-day images. Almost all the evaluation metrics for daytime and nighttime images are largely improved by our approach, which proves the superiority of our method.


\begin{table}
  \renewcommand\arraystretch{1.2}
\tabcolsep=0.15cm
 \caption{Quantitative comparison with state-of-the-art methods. The maximum depth is set to be 60m. Lower value is better for the first four metrics, higher value is better for the other three. The best results are presented in \textbf{bold} for each metric. Monodepth2 \cite{Godard_2019_ICCV} (day) and Monodepth2 \cite{Godard_2019_ICCV} (night) mean training with daytime and nighttime images of the Oxford dataset, respectively. Monodepth2+CycleGAN \cite{Zhu_2017_ICCV} means training the Monodepth2 model with daytime image, then using 'fake' daytime images, which is translated from the nighttime images, to test its performance. }

  \centering
  \begin{tabular}{c|ccccccc}

    \hline
    \rowcolor{blue!20}Method (test at night)     & AbsRel     & Sq Rel & RMSE& RMSE log  & $\delta$ <1.25 & $\delta$ <$1.25^2$ & $\delta$ <$1.25^3$\\
    \hline
    Monodepth2\cite{Godard_2019_ICCV} (day) & 0.440  & 6.039   & 14.148 &0.481&0.365&0.645&0.840  \\
    Monodepth2\cite{Godard_2019_ICCV} (night)& 0.513  & 8.266   & 16.668 &0.528&0.575&0.860&0.917 \\
    Monodepth2+CycleGAN\cite{Zhu_2017_ICCV}     & 0.240  & 2.341  & 9.082 &0.281&0.636&0.880&0.955 \\
    ADFA\cite{10.1007/978-3-030-58604-1_27}     & 0.233&3.783&10.089&0.319&0.668&0.844&0.924 \\
    RNW-Net \cite{Wang_2021_ICCV}    & 0.242&3.496&11.962&0.323&0.637&0.858&0.939\\
    ADDS-DepthNet \cite{Liu_2021_ICCV}     & 0.231&2.674&8.800&0.286&0.620&0.892&0.956 \\

    GlocalFuse-Depth   & \textbf{0.217}&\textbf{2.299}&\textbf{8.520}&\textbf{0.261}&\textbf{0.672}&\textbf{0.901}&\textbf{0.961}  \\

    \hline
     \rowcolor{blue!20}Method(test at day)    & AbsRel     & Sq Rel & RMSE& RMSE log  & $\delta$ <1.25 & $\delta$ <$1.25^2$ & $\delta$ <$1.25^3$\\
    \hline
    Monodepth2\cite{Godard_2019_ICCV} (day) &0.125&1.248&6.634&0.190&0.822&0.948&0.985  \\
    Monodepth2\cite{Godard_2019_ICCV} (night)     &0.305  &   3.648  &  11.592  &   0.396  &   0.459  &   0.745  &   0.885  \\
    ADDS-DepthNet \cite{Liu_2021_ICCV}     &0.115 & 0.794& 4.855&0.168&0.863&0.967&\textbf{0.989} \\
    GlocalFuse-Depth  &\textbf{0.113}&\textbf{0.785}&\textbf{4.787}&\textbf{0.165}&\textbf{0.871}&\textbf{0.969}&\textbf{0.989}  \\
    \hline
  \end{tabular}
\end{table}
\subsection{Qualitative Results}

\begin{figure}[htb]
\centering\includegraphics[width=16cm]{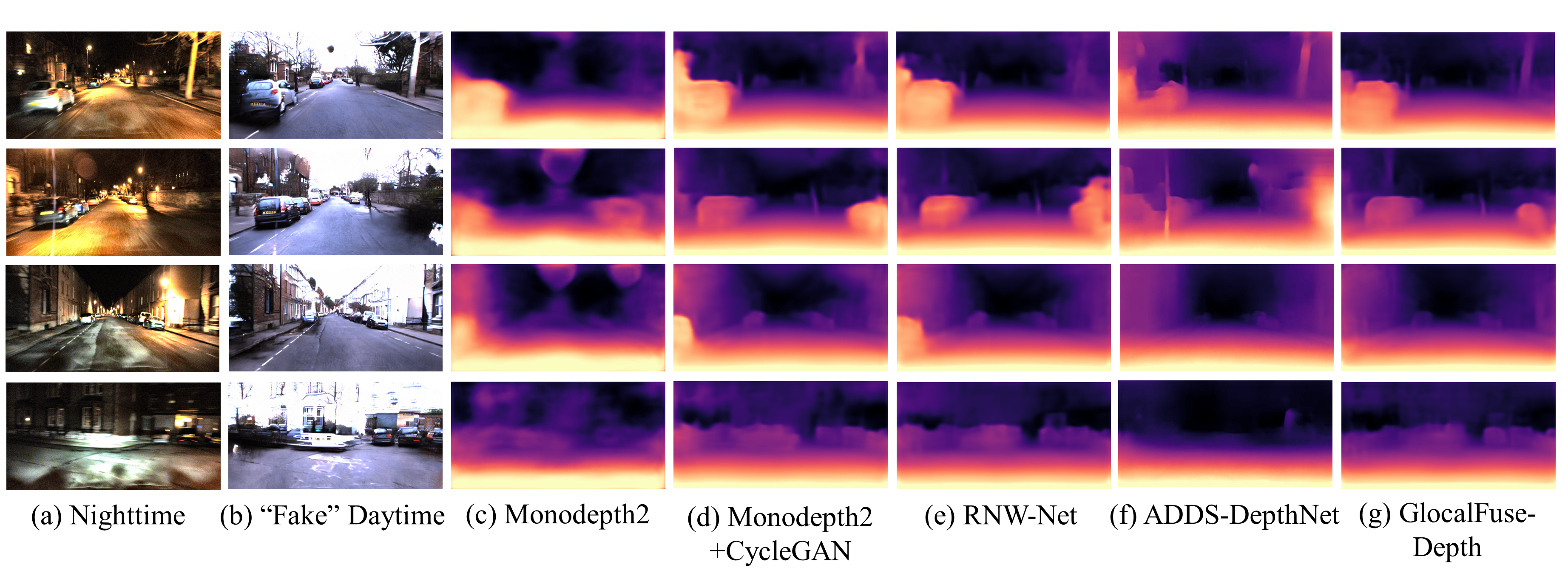}
\caption{Qualitative depth estimation result comparison with other state-of-the-art methods of nighttime images. From left to right: (a) Nighttime images, (b) 'Fake' daytime images generated by CycleGAN \cite{Zhu_2017_ICCV}, (c) Monodepth2 \cite{Godard_2019_ICCV}, (d) Monodepth2+CycleGAN\cite{Zhu_2017_ICCV}  , (e) RNW-Net \cite{Wang_2021_ICCV}, (f) ADDS-DepthNet \cite{Liu_2021_ICCV}, (g) GlocalFuse-Depth.}
\end{figure}

The qualitative nighttime image depth estimation results comparison of our method with some state-of-the-art methods are shown in Fig. 4, where (b) shows the CycleGAN \cite{Zhu_2017_ICCV} generated 'fake' daytime images  from nighttime images, (c) shows the depth estimation results of Monodepth2 \cite{Godard_2019_ICCV} trained with daytime images and tested with nighttime images, (d) shows the results of Monodepth2 \cite{Godard_2019_ICCV} trained with daytime images and tested with 'fake' daytime images. Compared with (c), it is obvious that (d) obtains better depth estimation results, which proves that the use of CycleGAN \cite{Zhu_2017_ICCV} could improve the performance of nighttime image depth estimation substantially. (e) shows the result of RNW \cite{Wang_2021_ICCV} and (f) shows the result of ADDS-DepthNet \cite{Liu_2021_ICCV}. (g) shows our result. Compared with (e) and (f), our result could obtain more accurate segmentation on some objects, such as the car on the left of the first and the third nighttime image. Besides, our method could be unaffected from some distortion of the nighttime image. In the second row, a defect line caused by image processing on nighttime images appears on ADDS-DepthNet \cite{Liu_2021_ICCV} estimation result. However, it disappears on our result, which proves the robustness of our method. Different from the first three row, which shows the scene of the car driving on a straightway, the nighttime image of the fourth row shows a rarely seen scene in training set: car turning. Our method could recover more details on the cars, while RNW \cite{Wang_2021_ICCV} and ADDS-DepthNet \cite{Liu_2021_ICCV} fails to do so, which proves the generalizability of our method. 

Fig. 5 demonstrates the qualitative daytime result comparison with other methods. We can see that more depth details can be recovered by our method, such as the tree on the left of the first daytime image and the car on the middle right of the second image, which clarifies the depth estimation of our method for daytime images.

\begin{figure}[htb]
\centering\includegraphics[width=16cm]{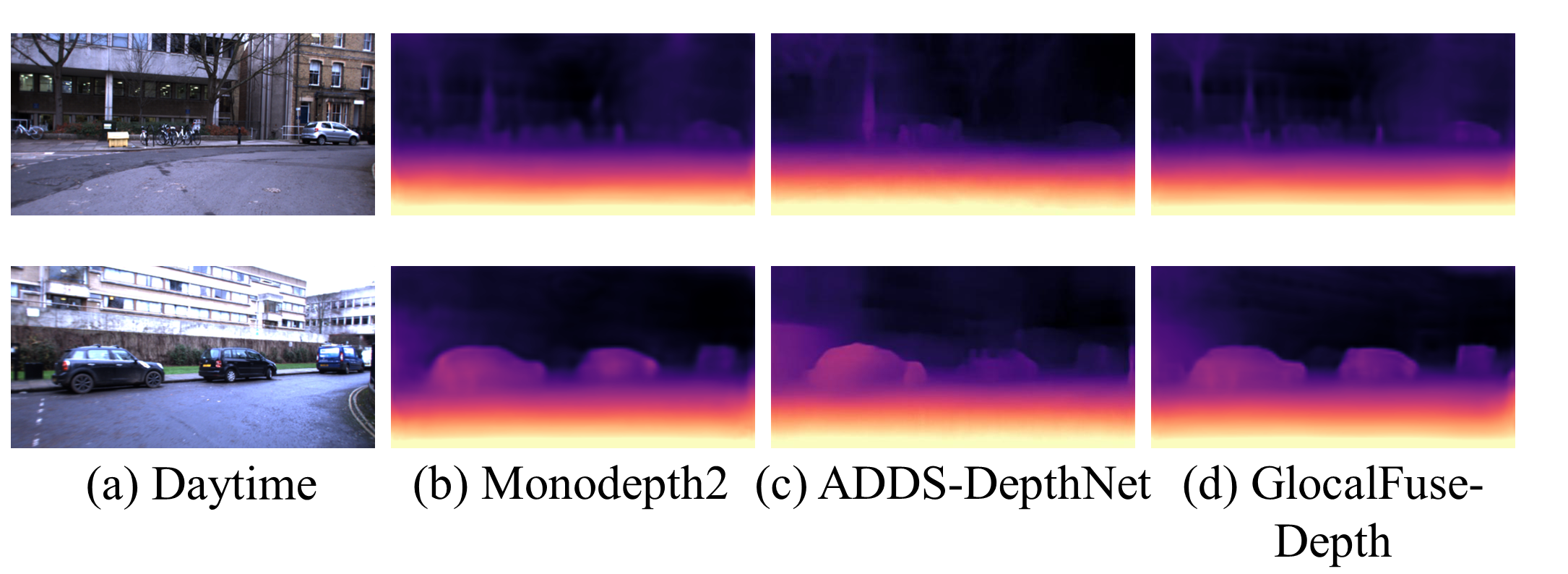}
\caption{Qualitative depth estimation result comparison with other state-of-the-art methods of daytime images. From left to right: (a) Daytime images, (b) Monodepth2 \cite{Godard_2019_ICCV}, (c) ADDS-DepthNet \cite{Liu_2021_ICCV}, (d) GlocalFuse-Depth.}
\end{figure}

\subsection{Ablation Study}
Here, we conduct a series of experiments to demonstrate the effectiveness of GlocalFuse-Depth and report the results in Table 2. For the sake of simplicity, we quantitatively compare the performance using AbsRel.

\begin{table}
\renewcommand\arraystretch{1.2}
\tabcolsep=0.15cm
 \caption{Ablation study on encoder design, composition of input image pair and fusion method.}
  \centering
  \begin{tabular}{c|c|ccccccc}
    \hline
     \rowcolor{blue!20}
     Study subject     & Method (night)&AbsRel     & Sq Rel & RMSE& RMSE log  & $\delta$ <1.25 & $\delta$ <$1.25^2$ & $\delta$ <$1.25^3$\\
     \hline
     &GlocalFuse-Depth&\textbf{0.217}&\textbf{2.299}&\textbf{8.520}&\textbf{0.261}&\textbf{0.672}&\textbf{0.901}&\textbf{0.961}\\ 
    \hline
    \multirow{2}{*}{Encoder design}&Transformer only&0.229  &   3.438  &  11.724  &   0.326  &   0.641  &   0.863  &   0.933 \\
    &CNN only &0.249  &   3.783  &  12.635  &   0.351  &   0.575  &   0.846  &   0.925 \\
    \hline
    \multirow{2}{*}{Image pair}&Day-day&0.232  & 3.626   & 11.853 &0.329&0.649&0.862&0.931\\
    &Night-night&  0.475&5.672&12.069&0.494&0.348&0.620&0.812  \\
    \hline
    \multirow{3}{*}{Fusion method}&Bifusion module\cite{10.1007/978-3-030-87193-2_2} &
     0.229&3.559&11.863&0.331&0.650&0.863&0.930\\
    &Concatenation &
     0.248&3.658&12.372&0.340&0.584&0.851&0.930\\
    &Dot product&
    0.286&3.573&12.149&0.339&0.502&0.809&0.950\\
    \hline 
     \rowcolor{blue!20}
    Study subject & Method (day)& AbsRel     & Sq Rel & RMSE& RMSE log  & $\delta$ <1.25 & $\delta$ <$1.25^2$ & $\delta$ <$1.25^3$\\
    \hline      &GlocalFuse-Depth&\textbf{0.113}&\textbf{0.785}&\textbf{4.787}&\textbf{0.165}&\textbf{0.871}&\textbf{0.969}&\textbf{0.989}  \\
    \hline  
    \multirow{2}{*}{Encoder design}&Transformer only &0.118  &   1.139  &   6.418  &   0.181  &   0.847  &   0.957  &   0.985 \\
    &CNN only & 0.134  &   1.404  &   7.181  &   0.206  &   0.805  &   0.940  &   0.980   \\
    \hline  
    \multirow{2}{*}{Image pair}&Day-day& 0.117 & 1.225   & 6.637 &0.186&0.839&0.951&0.983  \\
    &Night-night &   0.266  &   4.344  &  13.014  &   0.370  &   0.570  &   0.832  &   0.917 \\
    \hline  
    \multirow{3}{*}{Fusion method}&Bifusion module\cite{10.1007/978-3-030-87193-2_2}  &0.123&1.256&6.678&0.190&0.834&0.952&0.983\\
    &Concatenation &0.137&1.454&7.388&0.211&0.796&0.936&0.978\\
        &Dot product&0.218&2.285&8.933&0.276&0.649&0.879&0.955 \\
    \hline
  \end{tabular}
  \label{tab:table}
\end{table}
Firstly, we compare the depth estimation performance with different encoder design. In Table. 2, the quantitative depth estimation result of our proposed method (CNN-Transformer as encoders), CNN only as encoder in both branches and transformer only as encoder in both branches is shown. As can be seen, 5.2\% and 0.4\% improvement of AbsRel is achieved by using both CNN and transformer as encoders compared to using transformer only for nighttime and daytime images, respectively. Similarly, 12.9\% and 15.7\% improvement of AbsRel is achieved compared to using CNN only. These result proves the effectiveness of use complementary encoders in extracting local and global information from the images.

Secondly, we show the performance improvement on using day-night image pair as input. In Table. 2, the quantitative result of our proposed method (day-night image pair), using day-day image pair and using night-night image pair as input is shown. Compared with using day-day image pair, 6.5\% and 0.3\% performance improvement of AbsRel is achieved by our method for nighttime and daytime images. Besides, 54.3\% and 57.5\% improvement of AbsRel for nighttime and daytime images is achieved compared with using night-night image pair. The results above proves that the use of day-night image pair could help the network extract domain invariant information of both daytime and nighttime images, thus the estimation performance is improved.

Finally, we compare the performance our proposed fusion module with other fusion methods. In Table. 2, compared with the Bifusion module in \cite{10.1007/978-3-030-87193-2_2}, simply do the concatenation and dot product, our proposed fusion module achieves improvement of AbsRel for nighttime and daytime images by 5.2\% and 8.1\%, 12.5\% and 17.5\% and 24.1\% and 48.2\%, respectively. These results is owing to the use of effective channel-selection and spatial attention in our proposed module.

\section{Conclusion}
In this paper, we propose a two-branch network GlocalFuse-Depth which combines CNN and Transformer with late fusion for self-supervised monocular depth estimation of all-day images. The resulting architecture leverages the inherent characteristics of CNNs on modeling fine-grained feature and the capability of Transformers on modelling global context. A novel fusion module is proposed to selectively fuse multi-dimensional features from the two branches with channel-selection and spatial attention. Experiments results on the Oxford RobotCar dataset demonstrate that GlocalFuse-Depth achieves state-of-the-art results for all-day images.

\section*{Acknowledgments}
Research Grants Council of the Hong Kong Special Administrative Region of China (HKU 17210522, HKU C7074-21G, HKU 17205321, HKU 17200219, HKU 17209018, CityU T42-103/16-N) and Health@InnoHK program of the Innovation and Technology Commission of the Hong Kong SAR Government.

\bibliographystyle{unsrt}  
\bibliography{references}

\end{document}